\documentclass[]{article}
\usepackage{amsmath,amssymb,amsthm,multirow,multicol,graphicx,subfigure}
\usepackage[]{hyperref}
\hypersetup{pdftitle={Polyp Detection and Segmentation from Video Capsule Endoscopy: A Review}, colorlinks=true, citecolor=blue, linkcolor=red}
\graphicspath{ {./figures/} }
\begin{document}

\title{Polyp Detection and Segmentation from Video Capsule Endoscopy: A Review}

\author{V. B. Surya Prasath\footnote{Computational Imaging and VisAnalysis (CIVA) Lab, Department of Computer Science, University of Missouri-Columbia, MO 65211 USA. E-mail: prasaths@missouri.edu}}

\date{}
\maketitle
\begin{abstract}Video capsule endoscopy (VCE) is used widely nowadays for visualizing the gastrointestinal (GI) tract. Capsule endoscopy exams are prescribed usually as an additional monitoring mechanism and can help in identifying polyps, bleeding, etc. To analyze the large scale video data produced by VCE exams automatic image processing, computer vision, and learning algorithms are required.  Recently, automatic polyp detection algorithms have been proposed with various degrees of success. Though polyp detection in colonoscopy and other traditional endoscopy procedure based images is becoming a mature field, due to its unique imaging characteristics detecting polyps automatically in VCE is a hard problem. We review different polyp detection approaches for VCE imagery and provide systematic analysis with challenges faced by standard image processing and computer vision methods.

\end{abstract}

\section{Introduction}\label{intro}

Video capsule endoscopy (VCE) is an innovating diagnostic imaging modality in gastroenterology, which acquires digital photographs of the gastrointestinal (GI) tract using a swallowable miniature camera device with LED flash lights~\cite{IM00,WCEHandbook2014}. The capsule transmits images of the gastrointestinal tract to a portable recording device. The captured images are then analyzed by gastroenterologists, who locate and detect abnormal features such as polyps, lesions, bleeding etc and carry out diagnostic assessments. A typical capsule exam consists of more than 50,000 images, during  its operation time, which spans a duration of $8$ to $10$ hours. Hence, examining each image sequence produced by VCE is an extremely time consuming process. Clearly an efficient and accurate automatic detection procedure would relieve the  diagnosticians of the burden of analyzing a large number of images for each patient.

\begin{figure}[t]
	\centering
	\includegraphics[width=2.5cm]{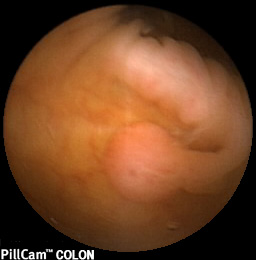}
	\includegraphics[width=2.5cm]{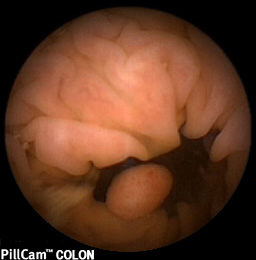}
	\includegraphics[width=2.5cm]{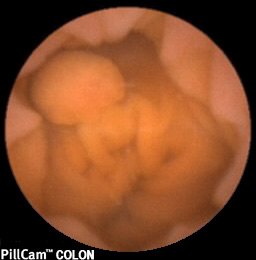}
	\includegraphics[width=2.5cm]{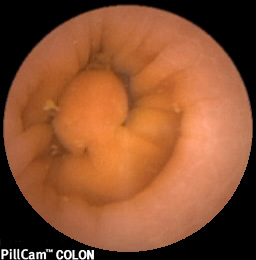}
	\includegraphics[width=2.5cm]{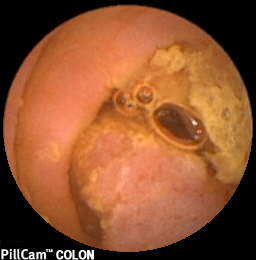}
	\includegraphics[width=2.5cm]{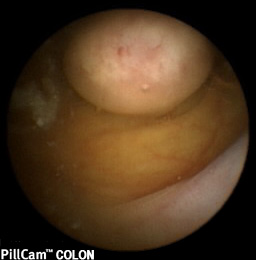}
	\includegraphics[width=2.5cm]{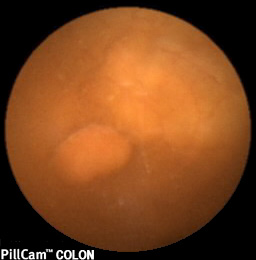}
	\includegraphics[width=2.5cm]{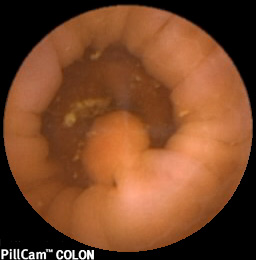}		
	\includegraphics[width=2.5cm]{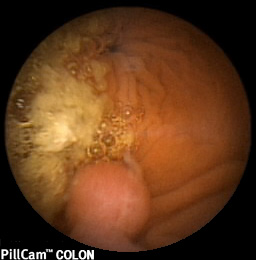}
	\includegraphics[width=2.5cm]{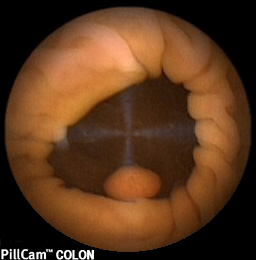}
	\includegraphics[width=2.5cm]{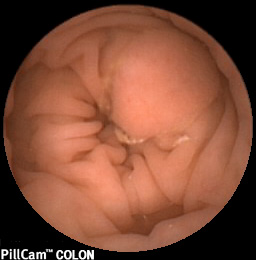}			
	\includegraphics[width=2.5cm]{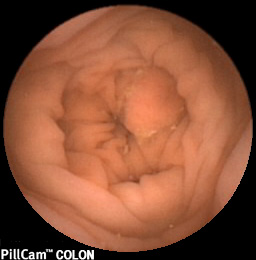}
	\includegraphics[width=2.5cm]{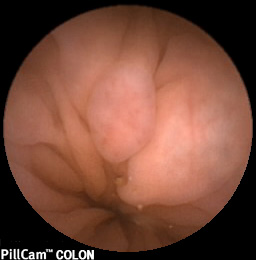}
	\includegraphics[width=2.5cm]{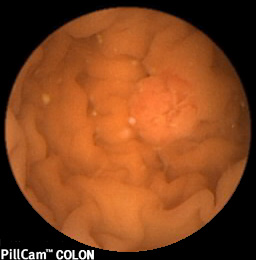}	
	\includegraphics[width=2.5cm]{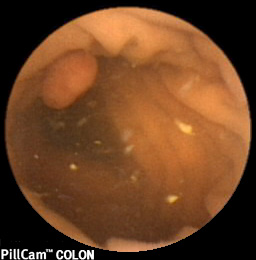}	
	\includegraphics[width=2.5cm]{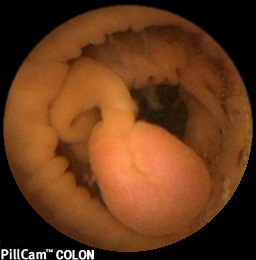}
	\includegraphics[width=2.5cm]{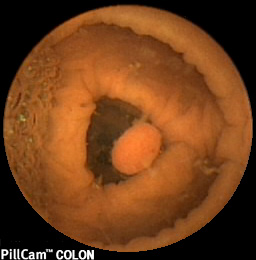}		
	\includegraphics[width=2.5cm]{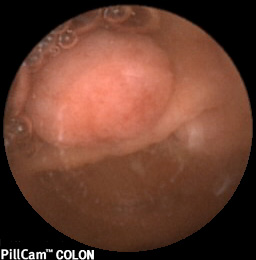}	
	\caption{Variability of appearance in colonic polyps under VCE imaging system. Images are taken from PillCam\textsuperscript{\textregistered} COLON capsule based exams on different patients. Notice the blurring, color, texture and geometric features (or lack of) at various levels. Other interference for feature detectors are turbid GI liquid, trash which are present due to VCE exams do not involve colon cleaning unlike traditional colonoscopy imaging techniques.}\label{fig:colon_polyps}
\end{figure}

Detecting polyps from VCE imagery is one of the foremost problems in devising an automated computer-aided detection and diagnosis systems. The polyps to be detected in the images are characterized by physicians according to human perception of their distinctive shapes, and also in some cases, by their color and texture on these geometric objects. In effect, according to medical information the geometry of colonic polyps  can be classified essentially in two types: pedunculate polyps, which are mushroom-like structures attached by a thin stalk to the colon mucosa, and sessile polyps, which are like caps (mushroom structures with no stalk). Their color is in general red, reddish or rose,  and their texture can be very similar to a human brain. Figure~\ref{fig:colon_polyps} shows example polyps across different regions of the gastrointestinal (GI) tract illustrating the variability in shape, color, and texture.  

Many previous studies have focused on detecting polyps on classical imaging technique of Colonoscopy (not capsule endoscopic images), as for example~\cite{iakovidis2005comparative,iakovidis2006intelligent,jiang2007novel,hwang2007polyp,park2012colon,bernal2012towards}. Polyp detection approaches in colonoscopy imagery include, using elliptical features~\cite{hwang2007polyp}, texture~\cite{iakovidis2005comparative,cheng2008colorectal}, color and position features~\cite{alexandre2008color,alexandre2007polyp}, see~\cite{ameling2009methods} for a review of polyp detection methods in Colonoscopy images. Polyp detection schemes applicable to colonoscopy and Computed Tomography (CT) colonography use mainly geometry based techniques, see for example~\cite{yoshida2001three}. However, due to the different imaging modality in VCE, images have different characteristics hence require unique methods for efficient polyp detection across various frames. Several shape based schemes were proposed to find polyps in virtual colonoscopy or computed tomography colonography and have been addressed; see e.g.~\cite{yoshida2001three,gokturk2001statistical,paik2004surface,konukoglu2007polyp,van2010detection,ruano2013shape}. Most of these methods take the already reconstructed surface representing the colon's interior or rely on some specific imaging techniques, see~\cite{liedlgruber2011computer,el2015automatic} for reviews. In contrast, VCE comes with an un-aided, uncontrolled photographic device, which moves automatically and is highly susceptible to illumination saturation due to near-field lighting~\cite{prasath2015automatic_contrast}. Moreover, the images from VCE differs significantly from images obtained with the traditional colonoscopy. For example, the liquid material in the lumen section is less in colonoscopy and hence the images look more specular. Whereas in VCE images the mucosa tissue looks diffusive under the presence of liquid and additionally the trash and turbidity can hinder the view of the mucosal surface~\cite{prasath2015automatic_contrast}. Due to the unaided movement of the capsule camera, blurring effects make the image looks less sharper. Moreover, the color of mucosal tissue under VCE has some peculiar characteristics~\cite{prasath2015fuzzification}. Due to these reasons particular to VCE its sensitivity for detecting colonic lesions is low compared with the use of optical colonoscopy as noted in~\cite{van2009capsule}. Nevertheless, a recent meta-analysis showed that capsule endoscopy is effective in detecting colorectal polyps~\cite{spada2010meta} (at-least in the colon capsules, though the jury is still out on the small-bowel and esophagus). Newer advances in sensors, camera system results in second generation capsule endoscopes and sensitivity and specificity for detecting colorectal polyps was improved~\cite{spada2011second,spada2012accuracy}. However, the increased imaging complexity and higher frame rates, though provide more information, inevitably puts more burden on the gastroenterologists. Thus, having efficient, robust automatic computer aided detection and segmentation of colorectal polyps is of great importance and need of the hour now. 

In this comprehensive survey paper, we provide an overview on different automatic image/video data based polyp detection (localization) and segmentation methods proposed in the literature so far (up-to September 2016\footnote{We refer the reader to the project website which is updated continuously with links to all the papers presented here and also to obtain more details about this research area:~\href{http://goo.gl/eAUWKJ}{http://goo.gl/eAUWKJ}}) and discuss the challenges that remain. We organized the rest of the paper as follows. Section~\ref{sec:contribution} provides a review of polyp detection, segmentation and holistic techniques from the literature. Section~\ref{sec:disc} we discuss the outlook in this field along with challenges that needs to be tackled by future research. 


\section{Review of polyp detection and segmentation in VCE}\label{sec:contribution}

Variable lighting and rare occurrence of polyps in a given (full) VCE video creates immense difficulties in devising a robust and data-driven methods for reliable detection and segmentation.  We can classify polyp detection/segmentation methods into two categories: (a) \textit{Polyp detection} - Finding where the polyp frame\footnote{Polyp frames may contain more than one polyp. We do not make a distinction of detecting one or multiple polyps in a given image.} occurs, not necessarily the location of the polyp within that frame (b) \textit{Polyp segmentation} - once a frame which contains polyp(s) is given segment mucosal area in which the polyp appears. Note that the first task is a much harder problem than the latter, due to the large number of frames the automatic algorithm needs to sift through to find the polyp frames which is usually a rare occurrence. Naturally, machine learning based approaches are essential for both categories, especially for polyp detection since the occurrence of polyp frames, frames where at least one polyp is visible, are very few in contrast to the typical full length of video frames (typically greater than 50000) in VCE imagery. The task is more complicated since colorectal polyps do not have common shape, texture, and color features even within a single patient's video. Nevertheless, there have been efforts in identifying polyp frames using automatic data-driven algorithms. For polyp segmentation, it is relatively an easier problem since the automatic algorithms need only to analyze a given polyp frame to find and localize the polyps that are present. We next review polyp detection and segmentation methods studied so far in the literature and discuss the key techniques used with relevant results.

\subsection{Polyp detection in capsule endoscopy videos}\label{sec:det}

\begin{figure}
	\centering
	\subfigure[Pedunculated/subpedunculated]{%
	\includegraphics[width=4.cm]{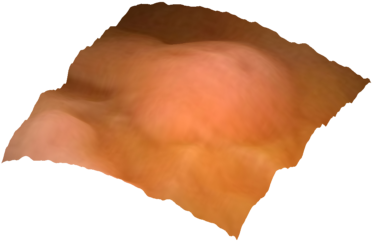}
	\includegraphics[width=4.cm]{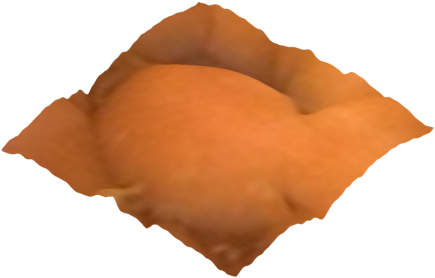}
	\includegraphics[width=4.cm]{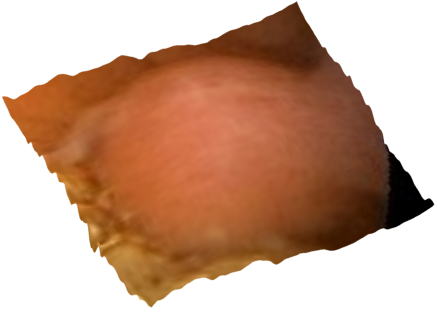}	
	\includegraphics[width=4.cm]{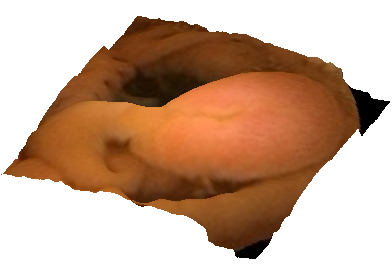}}
	\subfigure[Sessile]{%
	\includegraphics[width=4.cm]{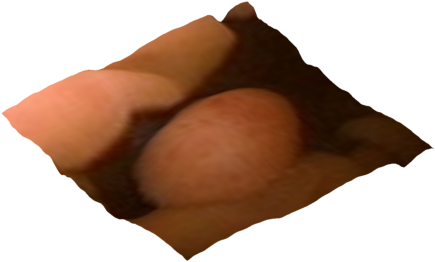}
	\includegraphics[width=4.cm]{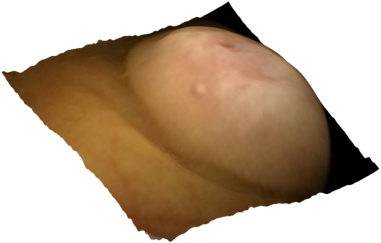}
	\includegraphics[width=4.cm]{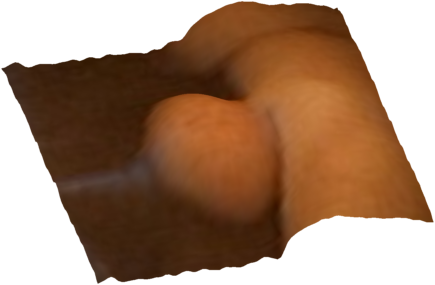}
	\includegraphics[width=4.cm]{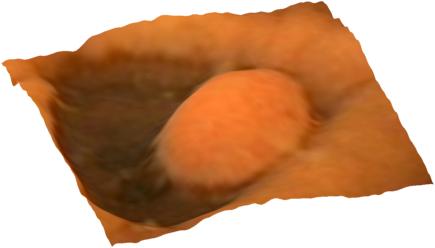}}
	\caption{Shapes or geometry of polyps significantly differ along the GI tract. Examples of polyps:
	(a) pedunculated/stalked or subpedunculated,
	(b) sessile. We show a 3D representation of VCE frames obtained using shape from shading technique~\cite{prasath2012mucosal}.}\label{fig:shape}
\end{figure}

\begin{figure}[t]
	\centering
	\includegraphics[width=2.5cm]{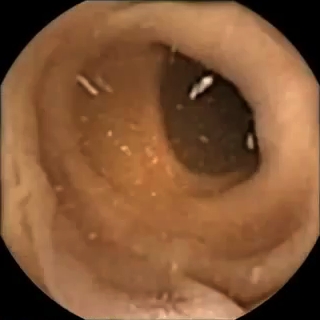}
	\includegraphics[width=2.5cm]{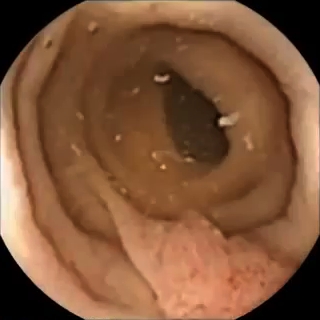}
	\includegraphics[width=2.5cm]{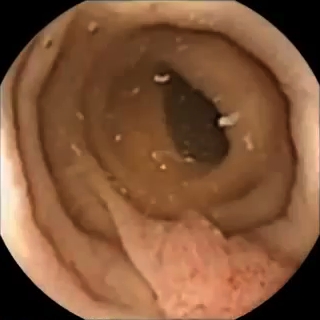}		
	\includegraphics[width=2.5cm]{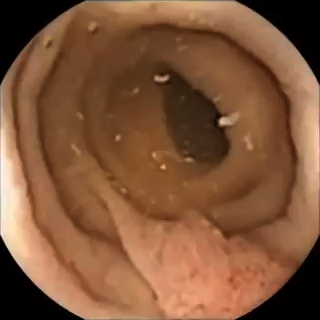}
	\includegraphics[width=2.5cm]{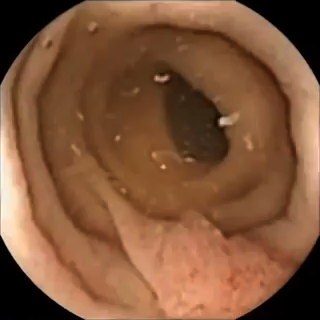}
	\includegraphics[width=2.5cm]{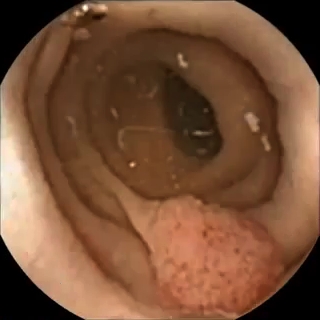}\\
	\includegraphics[width=2.5cm]{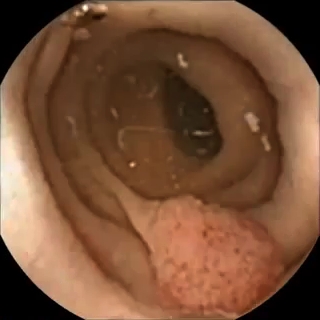}
	\includegraphics[width=2.5cm]{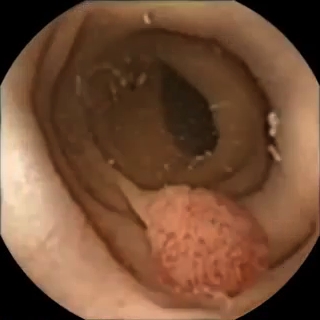}
	\includegraphics[width=2.5cm]{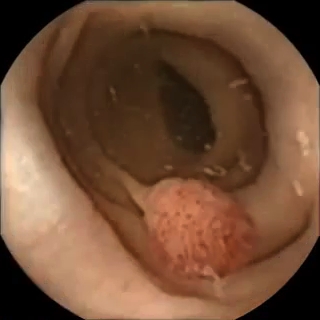}
	\includegraphics[width=2.5cm]{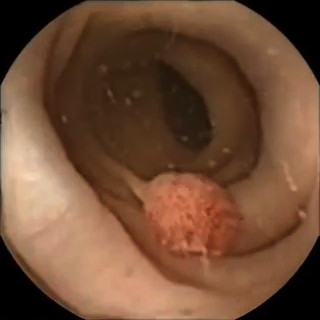}		
	\includegraphics[width=2.5cm]{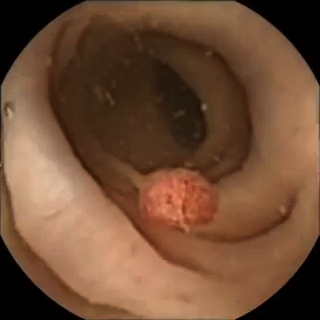}
	\includegraphics[width=2.5cm]{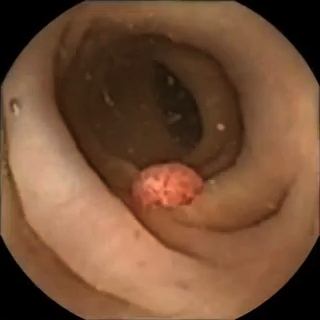}\\
	\includegraphics[width=2.5cm]{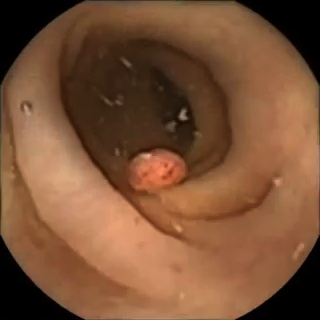}
	\includegraphics[width=2.5cm]{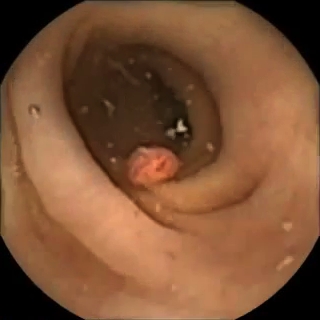}
	\includegraphics[width=2.5cm]{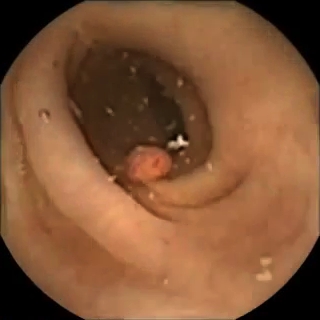}		
	\includegraphics[width=2.5cm]{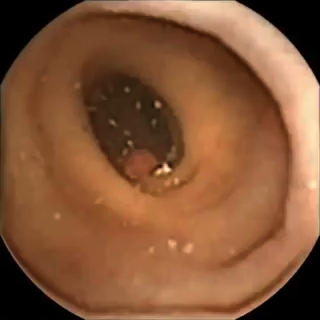}
	\includegraphics[width=2.5cm]{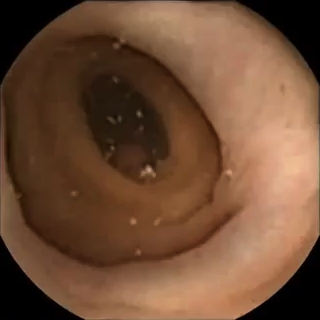}
	\includegraphics[width=2.5cm]{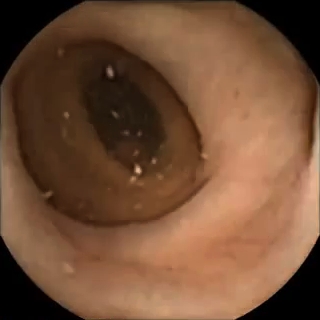}\\	
	\caption{Shapes of polyps vary across different frames, within a single patient VCE exam, due to the unconstrained movement of the camera. Top row: Data from PillCam\textsuperscript{\textregistered} COLON 2 polyp occurring in consecutive neighboring frames ($\#117$ to $122$). Bottom two rows: Other frames ($\#123$, $128$, $133$, $138$, $143$, $148$, and $153$,  $158$, $163$, $168$, $173$, $178$) from the polyp sequence where the shape and color undergoes changes.}\label{fig:polypseqs}
\end{figure}

There are two main classes of polyps with respect to their appearance in shape; pedunculated and sessile. In Figure~\ref{fig:shape} we show some example polyps (selected from images shown in Figure~\ref{fig:colon_polyps}) in 3D using the shape from shading technique~\cite{prasath2012mucosal}, indicating the amount of protrusion out of the mucosa surface\footnote{Visualizations as 3D figures are available in the Supplementary and also at the project website.}. Figure~\ref{fig:polypseqs} shows a pedunculated polyp which appears in consecutive frames from a VCE exam of a patient. The appearance changes drastically (rotation, translation, and scale changes), and the polyp which is not so clear in the first few frames (Figure~\ref{fig:polypseqs}(top row)) then becomes completely visible (Figure~\ref{fig:polypseqs}(middle row) and then becomes smaller before vanishing from the view. In the last few frames (Figure~\ref{fig:polypseqs}(bottom row)) the stalk of the polyp is not visible thus appearing like a sessile polyp attached to the mucosal fold at the bottom. Also note that the texture feature (due to vascularization) on top of the polyp is distinctively different from the surrounding mucosal folds which do contain small scale textures. 

One of the earliest works in VCE image processing and especially in detecting polyps automatically is by Kodogiannis et al~\cite{kodogiannis2007adaptive}, who studied an adaptive neuro-fuzzy approach. By utilizing texture spectrum from six channels (red-green-blue:RGB and hue-saturation-value:HSV color spaces) with adaptive fuzzy logic system based classifier they obtained $97\%$ sensitivity on $140$ images with $70$ polyp frames. 

Li et al~\cite{li2009comparative} compare two different shape features to discriminate polyp from normal regions. They utilize MPEG-7 shape descriptor (angular radial transform - ART), Zernike moments as features along with multi-layer perceptron (MLP) neural network as the classifier. Due to invariance to rotation, translation, and scale change Zernike moments are well suited to for polyp detection in VCE which has unconstrained movement of the camera.  They test their approach on $300$ representative images out of which $150$ contain polyps and achieved an accuracy of $86.1\%$. However, their approach only compares two specific shape features and discards color and texture information entirely. In a related work, Li et al~\cite{li2009intestinal} exploit combined color and shape features for polyp detection with HSI (hue, saturation, intensity) color space. They use compressed 2D chromaticity histogram (from hue and saturation), and Zernike moments (from intensity) and tested MLP and support vector machines (SVM) classifiers. On $300$ representative images out of which $150$ contain polyps, 2D chromaticity histogram plus 5th order Zernike moments with MLP obtained $94.20\%$ accuracy when compared to color wavelet covariance feature based result of $70.50\%$ with SVM. The chromaticity histogram is compressed with discrete cosine transformed and only lower frequency coefficients. This quantization scheme affects the performance with respect to color discriminability. 

Karargyris and Bourbakis~\cite{karargyris2009identification} performed Log-Gabor filter based segmentation along with SUSAN edge detector~\cite{smith1997susan}, curvature clusters, and active contour segmentation to identify polyp candidates. On a $50$-frame video containing $10$ polyp frames they achieved a sensitivity of $100\%$ (all 10 frames are selected as polyp candidates by the scheme). The method relies heavily on a geometric rule which assumes that the polyp region is an ellipse close to a circle which limits its applicability to detecting polyps of different shapes. Moreover, due to the small test cases of polyps it is not clear how the proposed approach performs in full VCE exam where the same polyp can have different geometric boundaries across frames.  Karargyris and Bourbakis~\cite{karargyris2011detection} extended later their previous work~\cite{karargyris2009identification} by adding SVM classifier. Hwang and Celebi~\cite{hwang2010polyp} used watershed segmentation with initial markers selected using Gabor texture features and K-means clustering that avoids the requirement of accurate edge detection considered in~\cite{karargyris2009identification}.On a set of $128$ images with $64$ polyp frames this method achieved a sensitivity of $100\%$.  However, a similar assumption to~\cite{karargyris2009identification} about the elliptical or circular shape is made and curvature clusters are used as an indicator for polyp candidates. 

Nawarathna et al~\cite{nawarathna2010abnormal} considered texton histograms for identifying abnormal regions with different classifiers such as SVM and K-nearest neighbors (K-NN). With Schmid filter bank based textons and SVM classifier an accuracy of $95.27\%$ was obtained for polyp detection.  
Nawarathna et al~\cite{nawarathna2014abnormal} later extended this approach with the addition of LBP feature. Further, a bigger filter bank (Leung-Malik) which includes Gaussian filters are advocated for capturing texture more effectively. Note that these approaches rely only texture features and do not include any color or geometrical features. The best results of $92\%$ accuracy was obtained for the Leung-Malik-LBP filter bank with K-NN classifier for $400$ images with $25$ polyps.

Figueiredo et al~\cite{figueiredo2011automatic} used a protrusion measure based on mean and Gaussian curvature for detecting polyps. They defined a novel protrusion measure which detected $80\%$ polyp frames accurately with localization of polyps. Unfortunately, not all polyps are protrusions (see Figure~\ref{fig:shape}(b) for examples), and the curvature index derived in~\cite{figueiredo2011automatic} fails to pick up VCE frames which contain sessile or flat polyps. Also note that this method does not rely on any classification and is purely a geometry based approach.

Zhao and Meng~\cite{zhao2011polyp} proposed to use opponent color moments with local binary patterns (LBP) texture feature computed over contourlet transformed image with SVM classifier with $97\%$ accuracy reported. Their work unfortunately do not mention how many total frames and polyp frames were used or how the color and texture features are fused. 

Zhao et al~\cite{zhao2011towards} studied a supervised classification approach with hidden Markov model (HMM) by integrating temporal information for polyp detection in VCE videos. They utilize a combination of color, edge and texture features which resulted in $118$-D feature vector which was reduced to $13$-D via Laplacian eigen-map method along with K-NN classifier. Their approach tested on $400$ images with $200$ polyp frames obtained an accuracy of $83.3\%$. However, the true strength of their method lies in testing image sequences, for this purpose they utilized  second dataset of $1120$ images ($224$ videos of $5$ frames duration) out of which there $560$ polyp frames. The obtained accuracy of $91.7\%$ for image sequences of $5$ frames length indicate strong promise of utilizing temporal information. In a related work, Zhao et al~\cite{zhao2012decision} used the same features along with a boosted classifier and evaluated using $1200$ VCE images with $90\%$ classification accuracy.

Hwang~\cite{hwang2011bag_polyp} used a bag of visual words model from computer vision literature by treating polyp regions as positive documents, and normal regions as the negative documents. The author used speeded up robust features (SURF) features quantize them with K-means clustering to generate the codebook to represent images as histogram of visual words. These feature vectors are then fed into SVM classifier and the results show $90\%$ sensitivity on a total of $120$ images with $60$ polyp frames. A very similar approach was undertaken in~\cite{hwang2011bag} with color features from HSI space added extra and tested on $250$ images with $50$ polyp frames and obtained $66\%$ sensitivity. 

Li and Meng~\cite{li2012automatic} applied the uniform LBP on discrete wavelet transformed sub-images to capture texture features on polyps. On a dataset of $1200$ images with $600$ polyp frames, an accuracy of $91.6\%$ was obtained with uniform LBP at $24$ circular set members and radius $3$ with SVM classifier. The method is based only on texture features and can fail to detect flat polyps with little texture, further there is no mention of the dimension of the feature set based on uniform LBPs which is usually high. 

Condessa and Bioucas-Dias~\cite{condessa2012segmentation} studied an extension of the protrusion measure in~\cite{figueiredo2011automatic}. By utilizing a two stage approach:  first stage involves multichannel segmentation, local polynomial approximation, and second stage extracts contour, curvature features with SVM classifier, they obtain $92.31\%$ sensitivity. The authors advocate using the temporal information via recursive stochastic filtering, though this has not yet been considered by researchers.

David et al~\cite{david2013automatic} utilized the protrusion measure from~\cite{figueiredo2011automatic}, however they observe that not all polyps exhibit high curvature values. By augmenting these high curvature peak based locations with color and illumination segmentation they apply a training stage with histogram of oriented Gaussians (HOG) features (computed on those locations) with MLP classifier. Results on $30540$ with $540$ polyp frames indicate an accuracy of $80\%$ is achieved with this approach. 

Figueiredo et al~\cite{figueiredo2013intelligent} combined the protrusion measure~\cite{figueiredo2011automatic} along with multiscale and uniform LBP texture features. They obtained an accuracy of $98.50\%$ on CIE Lab color space with monogenic LBP with linear SVM classifier on $400$ images with $200$ polyp frames. The protrusion measure alone achieved $65.5\%$ accuracy indicating that the reliance on polyps protrusion only is not a good assumption and further with an addition of basic LBP feature the accuracy increased to $97.25\%$ indicating the importance of texture. 

Yuan and Meng~\cite{yuan2014polyp} used SIFT feature vectors with K-means clustering for bag of features representation of polyps. By integrating histograms in both saliency and non-saliency regions the authors calculate weighted histogram of the visual words. These are fed into an SVM classifier and experiments on $872$ images with $436$ polyp frames show that $92\%$ detection accuracy was obtained. In a related work Yuan and Meng~\cite{yuan2014novel} tested Gabor filter and monogenic LBP features with SVM classifier. Their results indicate an accuracy of $91.43\%$ on the same dataset. 
Yuan et al~\cite{yuan2016improved} later extended these approaches with the addition of LBP, uniform LBP, complete LBP, HOG features for capturing texture information along with SIFT feature. They tested with SVM and  Fisher's linear discriminant analysis (FLDA) classifiers with different combinations of local features. Their method with SIFT + complete LBP with SVM classifier achieved top classification accuracy of $93.2\%$ on a set of $2500$ images which contained $500$ polyp frames. 

Mamonov et al~\cite{mamonov2014automated} proposed a system based on sphere fitting for VCE frames. The pipeline consist of considering the gray-scale image and applying a cartoon + texture decomposition. The texture part of the given frame is enhanced with nonlinear convolution and then mid pass filtered to obtain binary segments of possible polyp regions. Then a binary classifier uses best fit ball radius as an important decision parameter to decide whether there is a polyp present in the frame or not. This decision parameter is based on the assumption that the polyps are characterized as protrusions that are mostly round in shape, hence polyps which violate this do not get detected. The method tested on $18968$ frames with $230$ polyp frames obtained $81.25\%$ sensitivity per polyp. Here per polyp basis is computed as correct detection of at least one polyp frame in the corresponding sequence by the automatic algorithm. 

Jia et al~\cite{jia2014accurate} used geometrical features via ellipse fitting (area, ratio of major and minor axis), and multiscale rotation invariant LBP, HOG texture features. With support vector regression they reported $64.8\%$ true positive rate on a total number of $27984$ frames with $12984$ polyp frames. The huge number of polyp frames are obtained by perturbing original $541$ polyp frames to obtain $12443$ extra samples for training. 

Zhou et al~\cite{zhou2014polyp} utilized RGB averaging with variance for polyp localization and radius measurement of suspected protruding polyps using statistical region merging approach. On a total of $359$ this approach with SVM classifier obtained $75\%$ sensitivity.  Gueye et al~\cite{gueye2015automatic} used SIFT features and BoF method. On $800$ frames with $400$ polyp frames their SVM classifier obtained classification rate $61.83\%$, and on $400$ frames with $200$ polyp frames the rate increased to $98.25\%$. 

\begin{table}
\centering
	\begin{tabular}{@{}l|l|l|l}	
	\hline
	Ref.				&	Features/Technique & Classifier(s) &	Total number (polyps)  \\
	\hline
	\cite{kodogiannis2007adaptive} & Texture spectrum from RGB, HSV	& Neurofuzzy	& 140(70)	\\
	\cite{li2009comparative}	&	ART descriptor + Zernike moments	& MLP 		&	300(150)\\
	\cite{li2009intestinal}		&	Chromaticity histogram + Zernike moments 	& MLP, SVM 		&	300(150)	\\
	\cite{karargyris2009identification}	&	Log-Gabor filter + SUSAN edge detector	&	$\times$ &50(10)	\\  	
	\hline
	\cite{hwang2010polyp}		&	Gabor filters + Watershed segmentation 	& 	$\times$ &	128(64)\\ 
	\cite{nawarathna2010abnormal}	& Filter banks based texton histogram 	&	K-NN, SVM &  400(25) \\ 
	\hline
	\cite{figueiredo2011automatic}	&	Protrusion measure via curvatures &	$\times$	&	1700(10)	\\
	\cite{karargyris2011detection}&		Log-Gabor filter + SUSAN edge detector		&	SVM &		50(10)\\
	\cite{zhao2011polyp}		& Opponent color moments + LBP + LLE    & SVM &	  2 videos\\
	\cite{zhao2011towards}	&	Color + edge + texture + HMM & weak k-NN	&	400 (200), 1120(560) \\ 
	\cite{hwang2011bag_polyp} &  SURF features + BoW +  K-means 	&	SVM  & 120(60) \\						
	\cite{hwang2011bag}		   &	Color + Gabor filters + BoW +  K-means 	& SVM &	250 (50)\\		
	\hline
	\cite{zhao2012decision}	&	Color + edge + texture + HMM  &	Boosted SVM  & 1200(600) \\
	\cite{li2012automatic}	& Uniform LBP  + wavelet transform  & SVM 		&	1200(600)	\\	
	\cite{condessa2012segmentation}	&	Local polynomial approximation + geometry	  & SVM	&	3 videos (40)\\
	\hline	
	\cite{david2013automatic}	&	Geometry + color  + HoG   & MLP &	30540(540)\\
	\cite{figueiredo2013intelligent}	&	Geometry + color  + Monogenic LBP & SVM &		400(200) \\
	\hline	
	\cite{yuan2014polyp}		&	SIFT + Saliency + BoF	&	SVM	& 872(436)\\
	\cite{yuan2014novel}		&	Gabor filter + Monogenic LBP + LDA & SVM  &  872(436)\\
	\cite{mamonov2014automated}	&	Texture + midpass filtering + ellipse fitting  & Binary &  18968(230)\\
	\cite{nawarathna2014abnormal}	&	Texton histogram + LBP &		K-NN & 400(25)\\
	\cite{jia2014accurate}	& Geometry + LBP + HOG & Regression	&		27984 (12984)\\ 
	\cite{zhou2014polyp}		&	RGB + Variance + radius & SVM  	& 359\\
	\cite{gueye2015automatic}	&	SIFT  + BoF + K-means	&	SVM	&	800(400)\\
	\hline
	\cite{yuan2016improved}	&	SIFT + complete LBP + BoF	&  SVM	&	2500(500)\\
	\hline
	\end{tabular}
	\caption{Overview of automatic polyp detection methods in VCE with salient techniques, classifiers, and total number of frames in VCE videos with polyp frames.  Abbreviations: ART - angular radial transform, BoW - bag of words, BoF - bag of features, HMM - hidden Markov model, SIFT - scale invariant feature transform, HoG - histogram of oriented gradients, LLE - locally linear embedding, LBP - local binary patterns, MLP - multilayer perceptron, SVM - support vector machines,  LDA - linear discriminant analysis, $\times$ - no classifier used.}\label{table:detection}
\end{table}

In summary, there have been a few number of automatic polyp detection methods, though unfortunately the full dataset description is lacking in many of these published works which makes it hard for us to benchmark them. Table~\ref{table:detection} summarizes all the polyp detection methods covered so far with main techniques, classifiers utilized along with tested dataset details (whenever available).  It can be seen that the popular classifier is linear SVM with radial basis function as kernel due to its simplicity and ease of use. However, it is our belief that these majority of these methods either overfit or underfit as the proposed methods are tuned to obtain best possible detection accuracy results for their corresponding datasets. 
All the aforementioned methods highlight the accuracy of polyp detection based on how many polyp frames are detected out of all the input frames given. However, as mentioned by  Mamonov et al~\cite{mamonov2014automated} per polyp accuracy over per frame basis is very important. This is since detecting minimum a single polyp frame in a (typically) consecutive frames based sequences will suffice to alert the gastroenterologist/clinician. Figure~\ref{fig:polypseqs} shows different frames of a single patient VCE exam wherein a sequence of length $55$ contain a pedunculated polyp\footnote{Original video available in the Supplementary.}.  It is clear that detecting any one of the frames as polyp frame is enough as the gastroenterologist can inspect the neighboring frames manually.  This prospective automatic polyp alert system can reduce the burden on gastroenterologists as the number of frames to be inspected can be dramatically reduced, from a tens of thousands of frames to a few hundred possible polyp sequences.

\subsection{Polyp localization or segmentation within a VCE frame}\label{sec:seg}

\begin{figure}
	\centering
	Localization methods\\
	\subfigure[Log-Gabor filter~\cite{karargyris2009identification}]{\includegraphics[width=1.8cm, height=2cm]{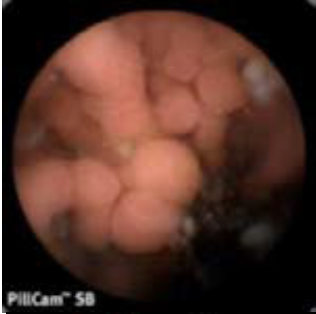}
	\includegraphics[width=1.8cm, height=2cm]{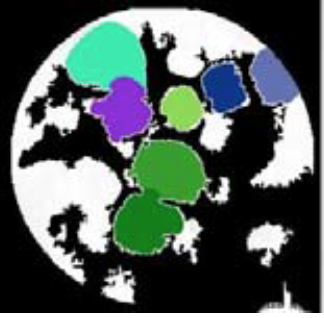}}
	\subfigure[K-means, curvature~\cite{hwang2010polyp}]{\includegraphics[width=1.8cm, height=2cm]{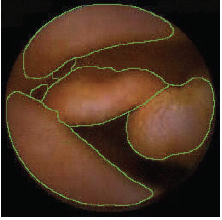}
	\includegraphics[width=1.8cm, height=2cm]{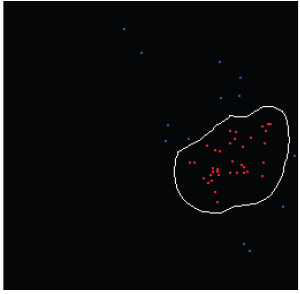}}
	\subfigure[Protrusion measure~\cite{figueiredo2011automatic}]{\includegraphics[width=1.8cm, height=2cm]{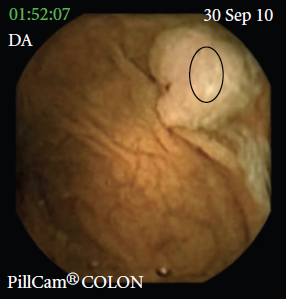}
	\includegraphics[width=1.8cm, height=2cm]{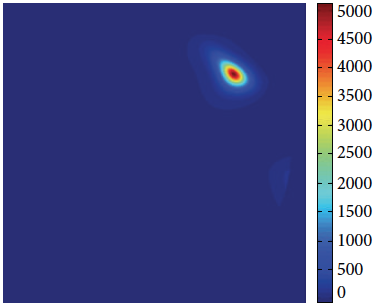}}
	\subfigure[Vascularization~\cite{prasath2015vascularization}]{\includegraphics[width=1.8cm, height=2cm]{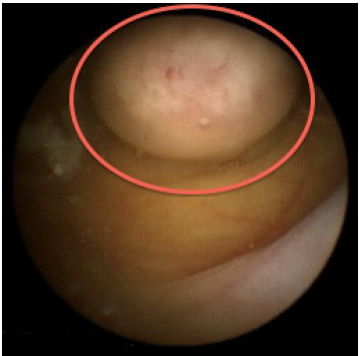}
	\includegraphics[width=1.8cm, height=2cm]{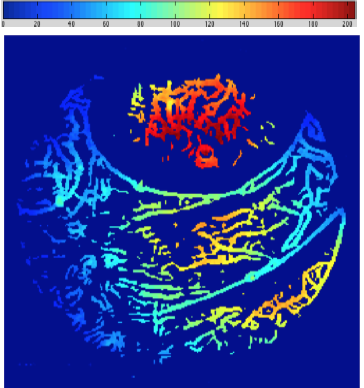}}\\
	\vspace{0.3cm}
	Accurate boundaries, segmentation\\ 
		\subfigure[Active contours~\cite{prasath2012mucosal}]{\includegraphics[width=1.8cm, height=2cm]{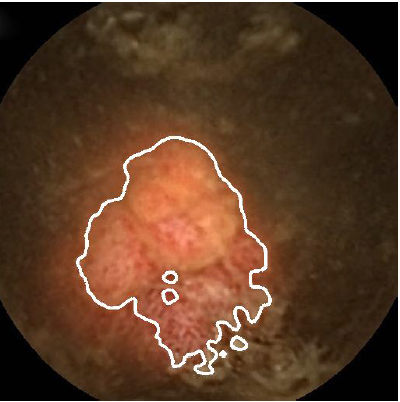}
	\includegraphics[width=1.8cm, height=2cm]{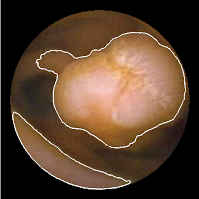}}
		\subfigure[Active contours~\cite{eskandari2012polyp}]{\includegraphics[width=1.8cm, height=2cm]{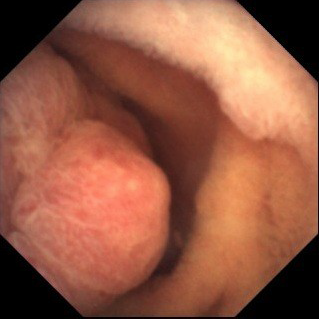}
	\includegraphics[width=1.8cm, height=2cm]{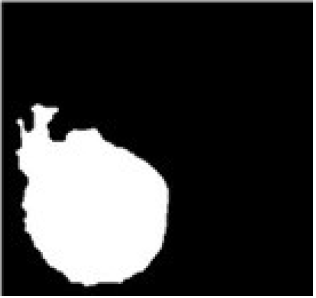}}	
		\subfigure[Active contours~\cite{alizadeh2014segmentation}]{\includegraphics[width=1.8cm, height=2cm]{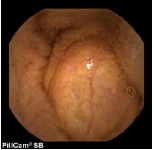}
	\includegraphics[width=1.8cm, height=2cm]{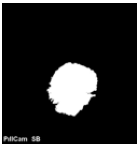}}
	\subfigure[$\alpha$ div active contours~\cite{meziou2014computer}]{\includegraphics[width=1.8cm, height=2cm]{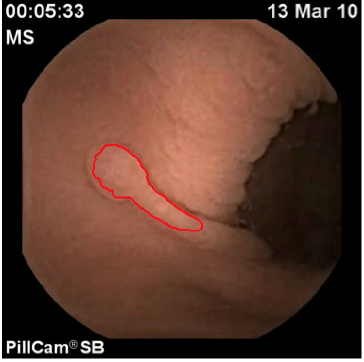}
	\includegraphics[width=1.8cm, height=2cm]{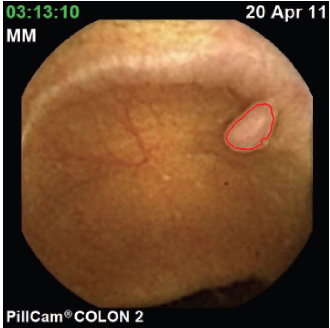}}
	\caption{Example polyps localization and segmentations. 
	Top row: Localization methods, 
	(a) Log-Gabor filter~\cite{karargyris2009identification},
	(b) K-means, curvature~\cite{hwang2010polyp},
	(c) protrusion measure~\cite{figueiredo2011automatic},
	(d) vascularization~\cite{prasath2015vascularization}.
	Bottom row: accurate boundaries and segmentation methods.
	(e)-(h) variants of active contours~	\cite{prasath2012mucosal,eskandari2012polyp,alizadeh2014segmentation,meziou2014computer}.}\label{fig:seg}
\end{figure}

Polyp localization or segmentation of polyps in a single frame of VCE is an (relatively easier) object identification problem. The general task falls under the category of image segmentation that is a well studied problem in various biomedical imaging domains. As we have seen before the color, texture and shape features individually are not discriminative enough to obtain polyp segmentation from VCE frames. There have been a number of efforts in polyp segmentation from VCE which we classify based on which segmentation techniques are utilized. Note that majority of the previously discussed polyp detection algorithms employ a polyp segmentation step to identify candidate polyp frames though \textit{accurately segmenting} the polyp region is not required for subsequent polyp detection in VCE. 

\begin{enumerate}

\item Localization

		\begin{itemize}

			\item Filtering: Karargyris and Bourbakis~\cite{karargyris2009identification} performed Log-Gabor filter based segmentation. Prasath and Kawanaka~\cite{prasath2015vascularization} proposed to use a novel vascularization feature which is based on Frangi vesselness filter.

			\item Geometry:  Hwang and Celebi~\cite{hwang2010polyp} derive a localization approach using curvature center ratio on K-mean clustering based segmented polyp frames. Figueiredo et al~\cite{figueiredo2011automatic} used a protrusion measure based on Gaussian and mean curvature to highlight possible polyp locations in a given VCE image.

			\item Hybrid: Jia~\cite{jia2015polyps} used K-means clustering and localizing region-based active contour segmentation. The results shown in this paper seems to be from colonoscopy imagery and not VCE. 

		\end{itemize}

\item Accurate boundaries, segmentation

		\begin{itemize}
			
			\item Active contours:  Meziou et al~\cite{meziou2012segmentation} used alpha divergence based active contours segmentation, though their approach is a general segmentation not for polyps in particular, see also~\cite{meziou2014computer}. Prasath et al~\cite{prasath2012mucosal} used active contours without edges method for identifying mucosal surface in conjunction with shape from shading technique, see also~\cite{PrasathMucosaCMNE11}.  Eskandari et al~\cite{eskandari2012polyp} used region based active contour model to segment the polyp region from a given image which contain a polyp, see also~\cite{alizadeh2014segmentation}.

	\end{itemize}
	
\end{enumerate}
Figure~\ref{fig:seg} highlights some example images and results obtained with the aforementioned automatic polyp localization, segmentation methods. Note that in localization approaches though polyp bounding boxes or circles were obtained there is no crisp boundary segmentations available. Obtaining accurate boundaries and segmentations of polyps in a frame requires a robust method which can avoid the pitfalls of bubbles, trash, illumination artifacts. A preliminary mucosa segmentation~\cite{PrasathMucosaCMNE11,prasath2012mucosal,prasath2015automatic,prasath2016automatic} maybe required before applying polyp segmentation step to avoid these non-polyp pixels. 

\subsection{Holistic systems}\label{ssec:holi}

\begin{table}
\centering
	\begin{tabular}{l|l|l|l}	
	\hline
	Ref.				&	Technique & Classifier(s) &	Total number (polyps)  \\
	\hline  
	\cite{romain2013towards}		&	Geometry + Texture & Boosting  &	1000(200) \\
	\cite{silva2013towards}	&	circular Hough + co-occurrence matrix & Boosting &	1500(300)\\ 
	\cite{silva2014toward}	&	Hough transform + co-occurrence matrix & Boosting &	1500(300)\\
	\hline
	\end{tabular}
	\caption{Holistic polyp detection and segmentation approaches for endoscopy systems. Note that all these proposed methodologies are so far only tested with traditional colonoscopy images~\cite{bernal2012towards}.}\label{table:holistic}
\end{table}

Romain et al~\cite{romain2013towards} studied a preliminary multimodal VCE for detecting polyps by utilizing active stereo vision within the capsule itself. It is based on two steps: (1) 2D identification polyp regions of interest (ROI) using simple geometric shape features, and (2) classification of polyps using 3D parameters computed on ROI using active stereo vision. Though, the authors of~\cite{romain2013towards} perform a prototype made of silicone for in vitro learning, the experiments were limited by the fact that there are no liquid (GI juices), trash and distinct lack of distension of the colon walls which affect the performance of automatic image analysis methods in real VCE exams. Similar efforts have been reported in~\cite{silva2013towards,silva2014toward}.
Table~\ref{table:holistic} summarizes these holistic system approaches proposed so far. Note that the polyp detection testing is done using a traditional colonoscopy images for now. Apart from these technological constraints, these approaches requires robust embeddable computer vision systems which needs to operate under strict energy budget (battery constraints). However, an computer vision embedded VCE system can revolutionize the diagnosis procedures which are done manually with tedious processes so far.

\section{Discussion and outlook}\label{sec:disc}

Automatic polyp detection and segmentation is nascent area which requires various computer vision methodologies such as geometrical primitives, color spaces, texture descriptors, features matching, along with strong machine learning components. 
A standard assumption made in classical colonoscopy imagery based polyp detection methods is that, polyps have high geometrical features such as well-defined shape~\cite{ruano2013shape} or protrusion out of mucosal surface~\cite{yoshida2001three}. Thus, curvature measures are widely utilized in detecting polyps and the adaptation to VCE imagery is undertaken with limited success~\cite{figueiredo2011automatic,condessa2012segmentation,david2013automatic,figueiredo2013intelligent}. 
Texture or color features based schemes such as~\cite{cheng2008colorectal,alexandre2008color} applied to colonoscopy images does not work well in WCE when used on their own as we have seen in Section~\ref{sec:det}. This is because, though the polyps appear to have different textural characteristics than the surrounding mucosa, bubble, trash and capsule motion induced by uncontrollable peristalsis have a strong effect on texture features. In such a scenario classifying with texture features alone is highly error prone. For example, the polyp shown in Figure~\ref{fig:polypseqs}(top row) has unique texture pattern, whereas for the polyp in Figure~\ref{fig:polypseqs}(middle row) texture appears to be uniformly spread across the mucosal structure surrounding it. Moreover, in a video, the neighboring frames show a partial view of the polyp with mucosa folds having a similar texture. 
Similarly, color of the polyp is not homogeneous across different polyp frames within a patient and highly variable across different exams from patients. 
In a comprehensive learning framework one can try to incorporate local and global texture features for discerning polyps along with vascularization, and color information. 
Existing approaches reviewed in Section~\ref{sec:contribution} are plagued by various factors. One of the main problem is the existence of trash, bubble since no colon cleaning is required in VCE exams. Robust polyp detection approaches must combine efficient trash, bubble detectors to avoid false positives. To conclude, we believe, a holistic view of combining, motion, geometry, color, and texture with a strong machine learning paradigm may prove to be a successful for robust, efficient automatic polyp detection in VCE imagery.

Based on the detailed description of the current state of the art polyp detection and segmentation methods we observe the following salient points for future outlook:

\begin{itemize}

	\item  Recent excitement generated by deep learning is very promising direction where massively trained neural network based classifiers can be used to better differentiate polyp frames from normal frames.  However deep learning networks in general require huge amount of training data, in particular labeled data of positive (polyp frames) and negative (normal frames) samples. One possible remedy for imbalanced data problem is to use data augmentation, therein one can increase the polyp frames by artificial perturbation (rotation, reflection,...), see e.g.~\cite{jia2014accurate} for an attempt to create a higher number of polyp frames for training. There have been some works in the last two years on endoscopy image analysis with deep learning~\cite{zhu2015lesion,cong2015deep,bae2015polyp}.

	\item Similar to ASU-Mayo Clinic polyp database for colonoscopy polyp detection benchmarking, VCE polyp detection requires a well-defined database with multiple expert gastroenterologists marked polyp regions. This will make the benchmarking and testing different methodologies for automatic polyp detection and segmentation standardized. 

	\item Sensor improvements with novel capsule systems~\cite{hatogai2012role,filip2011self} such as as more control in terms of higher image resolution, standardized illumination/contrast, controlled capsule speed, variable image capturing mechanisms  can help automatic image analysis. 
		
	\item Finally, embedding the image analysis part within capsule endoscopy imaging systems~\cite{silva2014toward,angermann2015smart} is an exciting research area which will enable the gastroenterologists can make real-time decisions. However, there are a lot of challenges remain for essential progress~\cite{iakovidis2015software}.
	
\end{itemize}

\section*{Acknowledgments}
This work was initiated while the author was a post doctoral fellow at the Department of Mathematics, University of Coimbra, Portugal under a UTAustin $|$ Portugal program and a Funda\c{c}\~{a}o para a Ci\^{e}ncia e a Tecnologia (FCT) fellowship. The author sincerely thanks Prof. Isabel N. Figueiredo (Mathematics, University of Coimbra, Portugal), Prof. Yen-Hsai Tsai (Mathematics, University of Texas, Austin, USA), and Dr. Pedro N. Figueiredo (Gastroenterology, University Hospital of Coimbra, Portugal) for initial collaborations in video capsule endoscopy image processing and analysis. 

\bibliographystyle{unsrt}
\bibliography{endoscopy_relatedrefs,endoscopy_polypsrefs}
\end{document}